\documentclass[letterpaper, 10 pt, conference]{ieeeconf}  

\IEEEoverridecommandlockouts                              

\usepackage{graphics} 
\usepackage{booktabs}
\usepackage{subfigure}
\usepackage{multirow}
\usepackage{epsfig} 
\usepackage{mathptmx} 
\usepackage{times} 
\usepackage{amsmath} 
\usepackage{amssymb}  
\usepackage{blindtext}
\usepackage{algorithmic}
\usepackage{textcomp}
\usepackage{xcolor}
\usepackage{threeparttable}
\usepackage{hyperref}
\usepackage[T1]{fontenc}
\usepackage{graphicx}
\usepackage{overpic}
\usepackage{makecell}
\usepackage{soul}

\title{Master Rules from Chaos:   \\ Learning to Reason, Plan, and Interact from Chaos for Tangram Assembly}

\author{Chao Zhao*, Chunli Jiang*, Lifan Luo, Guanlan Zhang, Hongyu Yu, Michael Yu Wang, Qifeng Chen
\thanks{\textsuperscript{*}Authors with equal contribution. To whom correspondence should be addressed czhaobb@connect.ust.hk, cqf@ust.hk. All Authors with the Hong Kong University of Science and Technology, Clear Water Bay, Hong Kong.}
} 

\usepackage{microtype}
\usepackage{tabularray}
\usepackage{gensymb}
\usepackage[top=0.75in, bottom=0.75in, left=0.75in, right=0.75in]{geometry}
\begin{document}
\maketitle
\thispagestyle{empty}
\pagestyle{empty}

\begin{abstract}

Tangram assembly, the art of human intelligence and manipulation dexterity, is a new challenge for robotics and reveals the limitations of state-of-the-arts. Here, we describe our initial exploration and highlight key problems in reasoning, planning, and manipulation for robotic tangram assembly. We present MRChaos (Master Rules from Chaos), a robust and general solution for learning assembly policies that can generalize to novel objects. In contrast to conventional methods based on prior geometric and kinematic models, MRChaos learns to assemble randomly generated objects through self-exploration in simulation without prior experience in assembling target objects. The reward signal is obtained from the visual observation change without manually designed models or annotations. MRChaos retains its robustness in assembling various novel tangram objects that have never been encountered during training, with only silhouette prompts. We show the potential of MRChaos in wider applications such as cutlery combinations. The presented work indicates that radical generalization in robotic assembly can be achieved by learning in much simpler domains. The code will be available \href{https://robotll.github.io/MasterRulesFromChaos/}{\color{blue}{https://robotll.github.io/MasterRulesFromChaos/}}.
\end{abstract}

\section{Introduction}

\begin{figure}[!t]
    \centering
    \begin{overpic}[width=\linewidth]{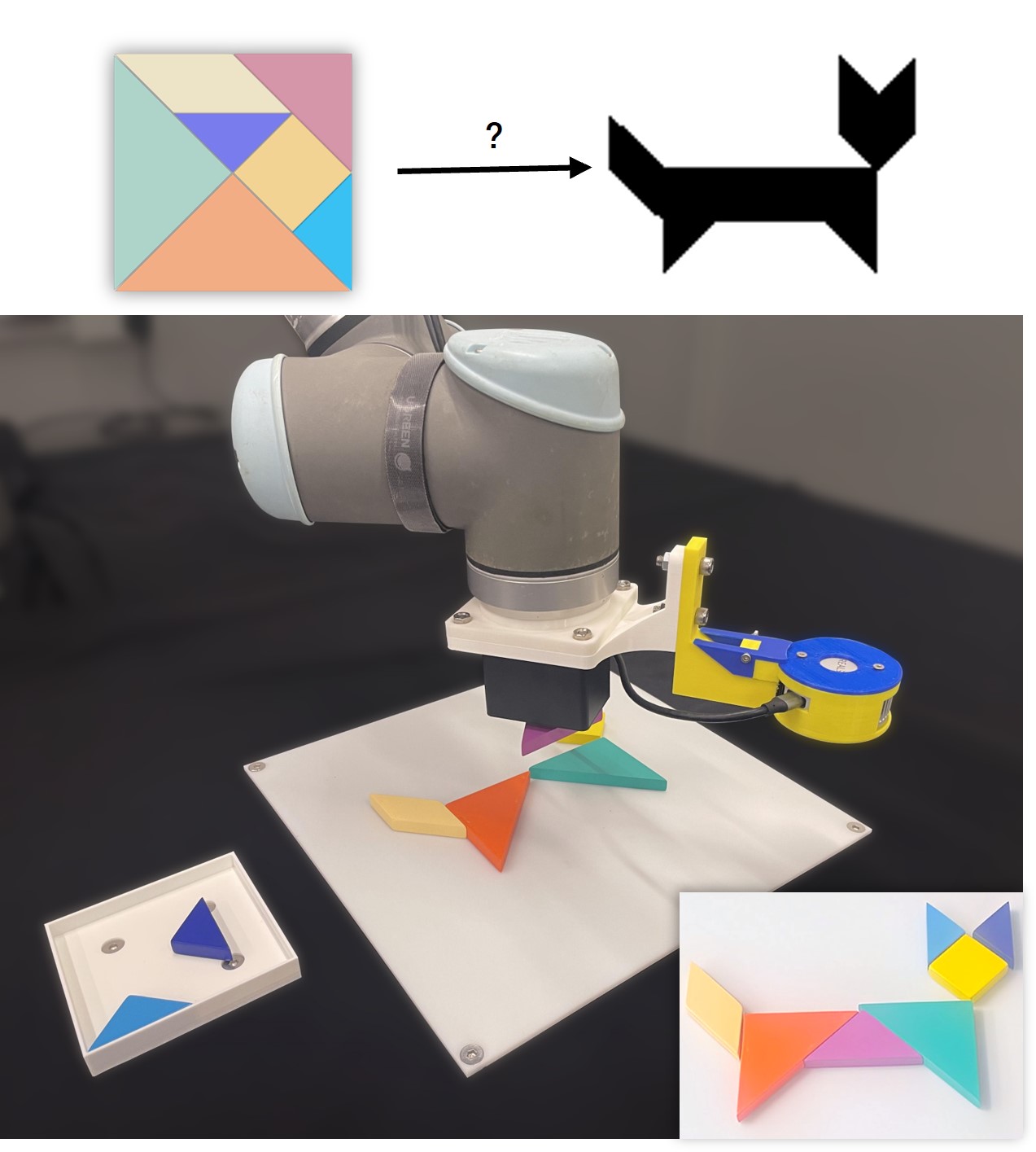}
    \put(16,97) {\small Pieces}
    \put(49,97) {\small Silhouette of Target Object}
    \end{overpic}
    \caption{The task of tangram assembly from the silhouette is to use seven pieces to assemble an object according to a silhouette prompt; The image on the bottom right shows the cat assembled by MRChaos.
    }
    \label{fig:fg1}
\vspace{-0.7cm}
\end{figure}

As robots are increasingly being utilized in society, robots are desired to assemble unfamiliar objects with appropriate reasoning ability, rather than conventional assembly tasks with high speed and high accuracy in well-defined environments such as production lines. In this regard, humans are far more skilled than robots. The clearest example is tangram assembly, an ancient asian puzzle game. The objective is to use seven pieces to assemble different objects in accordance with silhouette prompts. The term "silhouette" denotes the contour formed by the tangram pieces. For example, Fig. \ref{fig:fg1} shows an example silhouette of a fox.

Taking inspiration from the age-old tangram, we introduce the tangram assembly task into robotics, as shown in Fig. \ref{fig:fg1}. Within real-world robotics applications, silhouette prompts can serve as an intermediate representation derived from a high-level planning system, especially relevant in scenes where direct, detailed assembly instruction is unfeasible. Furthermore, in numerous applications, robots are presented with incomplete or ambiguous information and are required to make sense of their surroundings and tasks with the available data. A silhouette prompt encapsulates this ambiguity, thereby mimicking the genuine cognitive challenges robots face. Moreover, robotic tangram assembly cannot be effectively addressed by geometric solvers that ignore real physical interaction. 

The process of assembling tangrams from silhouette prompts necessitates a combination of robust reasoning, planning, and interaction skills, which are inherent to humans but still pose challenges for robots. The reasoning aspect involves deducing solutions from silhouettes without clear indicators like object poses or segmentation, making the correct piece placement difficult. Interaction challenges arise because, unlike conventional tasks with stable, constrained components, tangram pieces are movable and unstable, complicating precise localization and contact. Additionally, planning is difficult, as robots must take the imperfect interaction and reasoning results into consideration to plan globally and adjust their next step by observing the dynamic interaction results for successful assembly. 

To address the above challenges, we propose MRChaos, a general self-supervised solution for object assembly that learns through reinforcement learning (RL). At each time step, the policy predicts action based on current visual observation and the silhouette image to control the robot assembling the next piece part. Fig. \ref{fig:fg1} shows an assembly example. The system obtains rewards based on the visual difference captured by the camera before and after assembling each piece, eliminating the need for human demonstrations or annotations. Furthermore, MRChaos only explores assembling randomly generated objects during training instead of directly learning how to assemble target objects. This distinguishes MRChaos from other methods that typically require the manual design of target assembly models with prior assembly rules and physics built into them, requiring high expertise and substantial cost.

The primary contribution of this work is to introduce tangram assembly tasks into robotics and propose a new approach, MRChaos. We demonstrate the effectiveness of MRChaos through challenging tangram assembly tasks and daily tasks, such as cutlery combinations. Our results show dawning properties that the complexity of assembling objects can be tamed without resorting to rigid and time-consuming modeling methods or risky and costly experimentation in the real world. This is likely to be substantial in both existing and future deployed robotic assembly systems, as it allows for generalization to previously unseen objects without additional knowledge and sophisticated models, making robots more flexible and adaptable.

\section{Related Work}\label{sec:related_work}

\textbf{Robotic assembly:} Robotic assembly is a challenging area of robotic research. Conventional assembly tasks are typically performed in well-defined environments and have a wide range of tasks, including peg-in-hole \cite{zhang2022learning}, kit assembly \cite{zakka2020form2fit}, and nut-and-bolt assembly \cite{narang2022factory}. The classic methods \cite{park2020compliant} rely on precise geometric positioning \cite{wan2020planning}, collision-free motion planning \cite{chen2021blocks}, and force-controlled manipulation \cite{suarez2016framework} to achieve high precision, accuracy, and reliability. However, these methods are sensitive to variations in the environment and tasks and require costly parameter tuning \cite{li2019survey}. Recently, some works have approached the problem of robotic assembly as pose estimation \cite{collet2011moped} or learning to shape correspondences directly \cite{zakka2020form2fit}. Another line of work has introduced RL to robotic assembly and has shown promising results \cite{ app10196923, 8793506, zhang2022learning, narang2022factory}. For example, \cite{inoue2017deep} utilized RL to learn insertion policies, while \cite{vuong2021learning} introduced manipulation primitives to decrease the exploration difficulty. 

Conventional assembly methods have proven successful in highly structured settings but are limited by factors such as the requirement for complete knowledge of all objects or substantial task-specific training data \cite{8957300}. Therefore, having policies robustly for novel object assembly has remained a grand challenge in robotics. This motivates us to propose MRChaos under the context of tangram assembly, which focuses on exploring how to learn policies that are able to generalize to new object assembly without relying on exhaustive modeling procedures or costly real-world data collection.

\textbf{Object arrangement:} An area close to robot assembly is object arrangement, which focuses on the positioning and organization of objects. To manipulate objects, non-prehensile methods such as shearing \cite{asencio2017shearing} and pushing \cite{song2020multi} have been developed. For planning, \cite{labbe2020monte} proposed using the Monte-Carlo tree search to find paths and actions. \cite{zeng2021transporter} argue that incorporating spatial symmetries can simplify the planning process. More recently, \cite{kapelyukh2022dall} and \cite{liu2022structformer} indicated that transforming language commands into high-level planners could be a promising approach. In contrast to object arrangement, robotic assembly focuses on integrating individual parts into a complete object, which involves contact-rich interactions. In particular, in tangram assembly, there is a delicate balance between the tight fitting of parts and preventing collision with the assembled pieces. 

\textbf{Tangram in robotics:} In the field of human-robot collaboration/interaction, studies have investigated the development of shared strategies for playing tangram puzzles between a human and a robot \cite{6309474}, exploring the use of tangram puzzles to examine artificial curiosity \cite{park2017growing} and the growth mindset \cite{rosenberg2021expressive}. Furthermore, recent studies have utilized tangram pieces to demonstrate grasping performance \cite{ai2020013} or simulate industrial products to assess production lines \cite{qin2022robot}. All the above works use only a single tangram piece or have full knowledge of the solution. In this paper, we introduce the tangram into the field of robotics assembly and focus on the challenges posed by assembling novel tangram objects from silhouettes.


\begin{figure*}[!t]
    \centering
    \begin{overpic}[width=\linewidth]{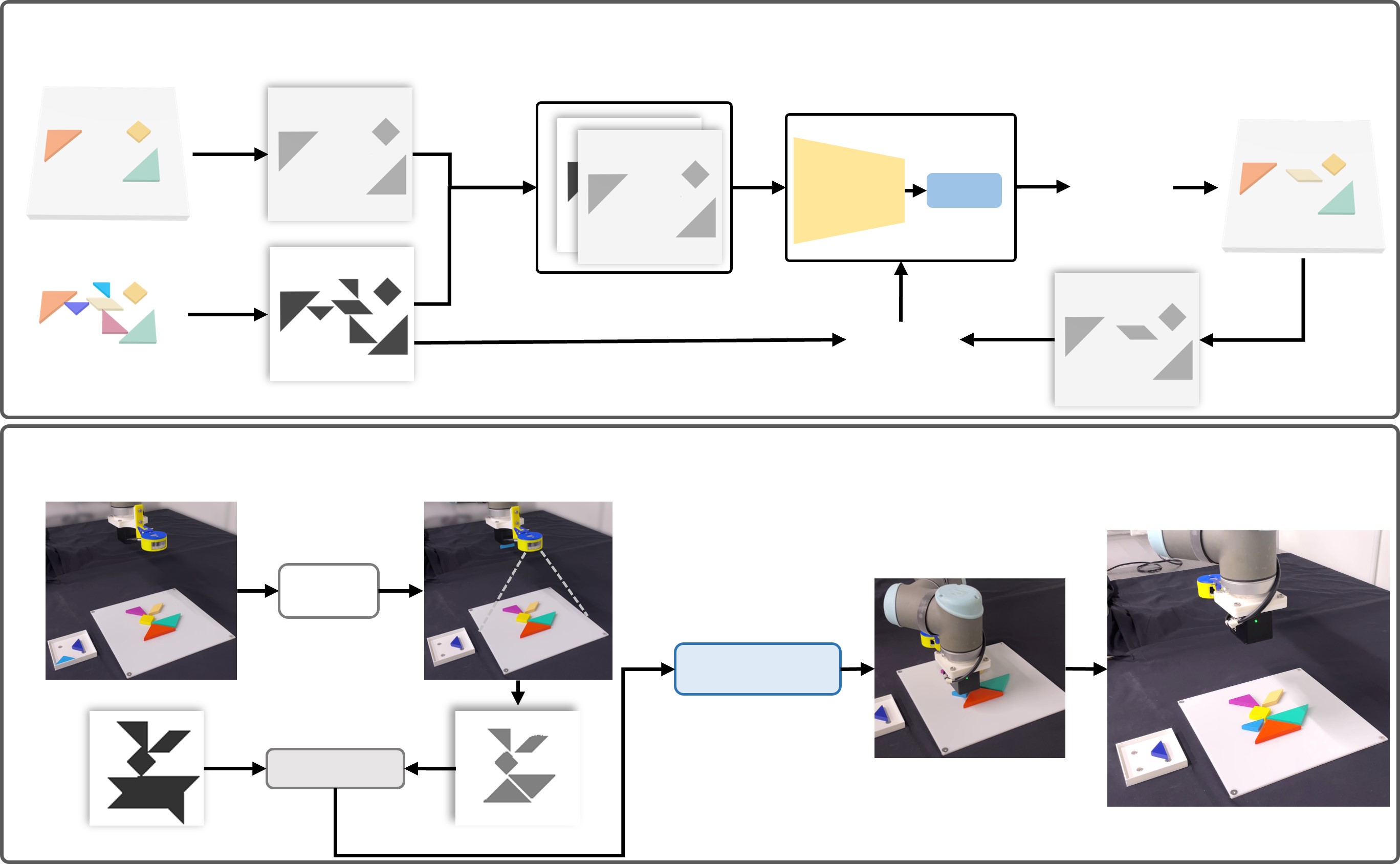}
    \put(1,59.25) {\small \textbf{A:} Learning assembly policy in simulation}
    \put(4.5,56.25) {\footnotesize Simulation}
    \put(18.65,56.25) {\footnotesize Depth observation}
    \put(4,35) {\footnotesize Generated}
    \put(3.25,33.5) {\footnotesize target object}
    \put(18.5,32.75) {\footnotesize Silhouette prompt}
    \put(30,51.25) {\footnotesize $I_{tc}$}
    \put(30,40.5) {\footnotesize $I_{s}$}
    \put(32.75,46.75) {\footnotesize Concat}
    \put(41,55) {\footnotesize Observation $o_{t}$}
    \put(59.5,52) {\footnotesize Policy network}
    \put(58.5,48.75) {\footnotesize Conv}
    \put(58,46.75) {\footnotesize encoder}
    \put(67.2,47.5) {\footnotesize MLP}
    \put(64.75,39.5) {\footnotesize Train with PPO}
    \put(77,47.9) {\footnotesize Action $a_{t}$}
    \put(61.2,37) {\footnotesize Reward $r_{t}$}
    \put(90.5,53.8) {\footnotesize Place $n_{t}$}
    
    \put(1,29) {\small \textbf{B:} Deployment in real-world}
    \put(8,26.5) {\footnotesize Step $t$}
    \put(21,19.75) {\footnotesize Suction}
    \put(21.1,18.25) {\footnotesize module}
    \put(34,26.5) {\footnotesize Grasp $n_{t}$}
    \put(2,1.3) {\footnotesize Prompts of novel objects}
    \put(16,7.5) {\footnotesize $I_{s}$}
    \put(30,7.5) {\footnotesize $I_{tc}$}
    \put(45.5,14.5) {\footnotesize $o_{t}$}
    \put(20,6.3) {\footnotesize Concatenate}
    \put(31,1.3) {\footnotesize Depth observation}
    \put(49.3,13.5) {\footnotesize Policy network}
    \put(66.5,21) {\footnotesize Action $a_{t}$}
    \end{overpic}
    \caption{\textbf{System Overview. A:} During training, a random target object, along with a silhouette image $I_s$, is generated. At time step $t$, the $I_{tc}$ is captured from a top-down camera and concatenated with $I_s$ as $o_t$. The agent receives the $o_t$ and outputs the action $a_t$ for the robot to place the $n_t$ piece. The agent finally receives rewards $r_t$ according to the visual difference and updates the policy with PPO; \textbf{B:} Deployment of MRChaos in the real world.
    }
    \label{fig:fg3}
\vspace{-0.65cm}
\end{figure*}
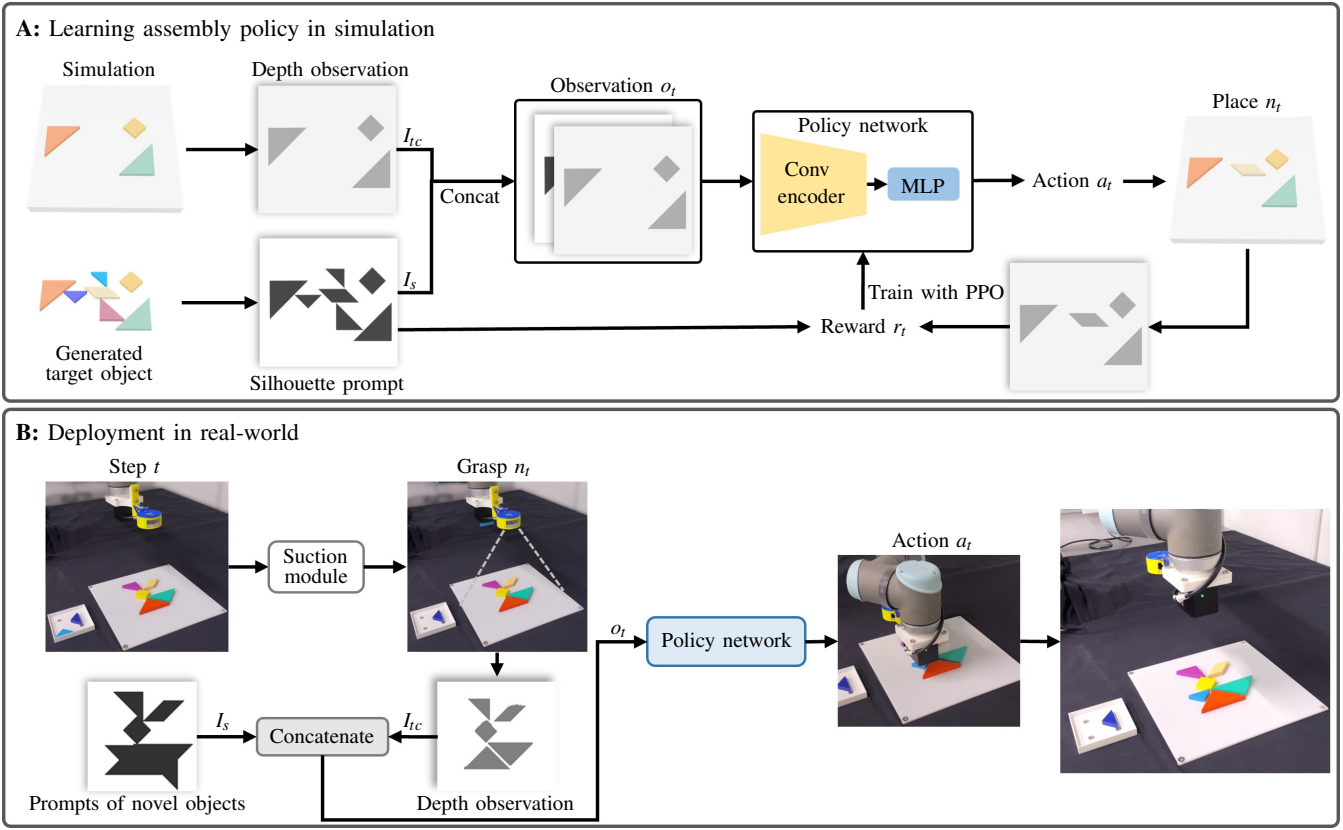

\section{System and Task Setup}

Our system setup consists of the following major components: a Universal Robot arm equipped with a suction gripper and a wrist-mounted Intel Realsense L515 depth camera (see in the video). We also create similar ones in the simulation. 

\subsection{Task setup}

We consider the tangram assembly task, which involves using seven pieces $N = (n_1, ..., n_j)$ to assemble an object. The $j$ equals seven for the tangram task but can be extended to other tasks with different numbers of pieces. The robot sequentially grasps each piece, starting with the largest and ending with the smallest. When arranged properly, these pieces can be assembled into complex objects with abstract semantics, such as animals and sports. The goal for the robot is to assemble the object using all pieces based on a silhouette image representing the target object.

\subsection{Problem formulation}
We formulate the tangram assembly problem as a Markov decision process (MDP) in discrete time dynamics. Formally, an MDP is composed of four components: a state space $S^{\prime}$, an action space $A$, a function of reward $R(s_t, s_{t+1})$, and a transition probability $P(s_{t+1} |s_t , a_t)$. At time step $t$, the agent predicts an action $a_t$ according to current policy $\pi(a_t |s_t)$ and observes the environment via $o_t$, the robot executes the action $a_t$ to place the $t^{th}$ tangram piece $n_t$ and obtains the reward $r_t$ from the environment. The goal of the agent is to find an optimal policy $\pi^*$ that maximizes the discounted sum of rewards over a finite time horizon.

\section{Method}\label{sec: method}

We propose MRChaos, a novel self-supervised solution that learns through RL. MRChaos aims to enable robots to learn policies for assembling novel objects with only silhouette prompts. Our key insight is that by learning to assemble randomly generated objects in simulation through self-exploration, the robot can develop strong reasoning, planning, and interaction abilities that enable it to assemble novel objects without prior experience. In the following section, we describe our method in the context of tangram assembly. We also present the required modifications for daily tasks in Sec. \ref{sec:exp}D. 

MRChaos consists of two stages, as shown in Fig. \ref{fig:fg3}. First, a policy is trained with RL to assemble randomly generated tangram objects in the simulation. These target objects are simple in composition, easy to assemble, and have no human-understandable semantic information. The reward is constructed by the coverage of the placed tangram part in the corresponding area of the target silhouette. Then, we transfer the learned policy to the robot in the real world. 

\subsection{Random Object Generation}

During training, we aim to find an optimal policy that enables the robot to successfully assemble tangram objects that are generated randomly. We model the policy as a discrete-domain stochastic policy and train using proximal policy optimization \cite{schulman2017proximal} (PPO). We use Pybullet as our simulator to build the training environment. Within the simulator, we generate target objects with random construction for learning assembly at each episode in two ways. The first is to construct new objects by randomly placing each piece while ensuring that there is no stacking. The second is to build on the first by placing a gravitational force that attracts the pieces at one point to encourage more bonding of the pieces together. 

\subsection{Observation and Action}

The observation is defined as $o_t=(I_{s}, I_{tc})$, where $I_{s}$ refers to the image of the silhouette prompt, and $I_{tc}$ refers to the visual observation of workspace at time step $t$, as shown in Fig.~\ref{fig:fg3}A. Notably, we do not specify the current time step explicitly in the observation, which requires the agent to reason based on the observation of the environment, reflecting strong reasoning ability. 

After each time step $t$, the robot picks the $t^{th}$ piece $n_t$ from its initial position and predicts the action to place the piece based on observations $o_t$. The action $a_t$ includes a gripper displacement, denoted as $(x_t, y_t, \theta_t)$. The displacement of the gripper refers to the relative variation between the pose after picking the piece and the desired one.



\subsection{Policy Architecture}

The policy network $\pi(a_t |s_t)$ is composed of a convolutional (Conv) encoder and a multilayer perceptron (MLP) as shown in Fig. \ref{fig:fg3}A. The $I_{s}$ and $I_{tc}$ are first concatenated into a $2\times120\times120$ vector as the input and encoded into a latent vector using a Conv encoder. The Conv encoder is constructed with three convolutional layers. The latent vector is then fed into subsequent MLP with 512 hidden units to compress inputs into a more compact representation and predict the action $a_t$. 

\subsection{Reward}

The reward is constructed by the visual difference between the silhouette image and the observation of the object assembled by the robot. More specifically, we calculate the coverage of the $n_t$ block in the silhouette after the robot places it at the time step $t$ and provides positive rewards based on coverage. Formally, the reward $r_t$ is defined as
\begin{equation}
r_t= \frac{|p_t\cap p_t^{\prime}|
}{|p_{t}^{\prime}|},
\end{equation}
where $p_{t}^{\prime}$ is the pixel set of piece $n_t$ on the silhouette image $I_s$, $p_{t}$ is the pixel set of piece $n_t$ on the image $I_{tc}$, $|p_t\cap p_t^{\prime}|$ is the number of overlapping pixels of $p_{t}^{\prime}$ and $p_{t}$, and $|p_{t}^{\prime}|$ is the pixel number in $p_{t}^{\prime}$. Since all states are available in the simulation, obtaining both sets $p_{t}^{\prime}$ and $p_{t}$ is a breeze. Thus, $r_t$ is between 0 and 1, denoting the similarity between the outcome of the tangram piece placement according to the predicted action and the target.

\subsection{Training Curriculum}

We propose a two-stage curriculum that aims to facilitate policy learning. During the first stage, a predetermined number of pieces, randomly ranging from $0$ to $j-1$, are pre-assembled, leaving only the task of assembling the next piece for the agent. In the second stage, the agent is tasked with assembling all pieces, starting when the rewards of the first stage cease to increase. The first stage simplifies the assembly task, allowing the agent to learn the assembly of local parts quickly. In the second stage, through continuous physical interaction and closed-loop visual feedback, the agent is encouraged to improve the strategy from the global perspective of the assembly task, avoiding prediction failures caused by the accumulation of minor errors at each step. 

\begin{figure}[!t]
    \centering
    \begin{overpic}[width=\linewidth]{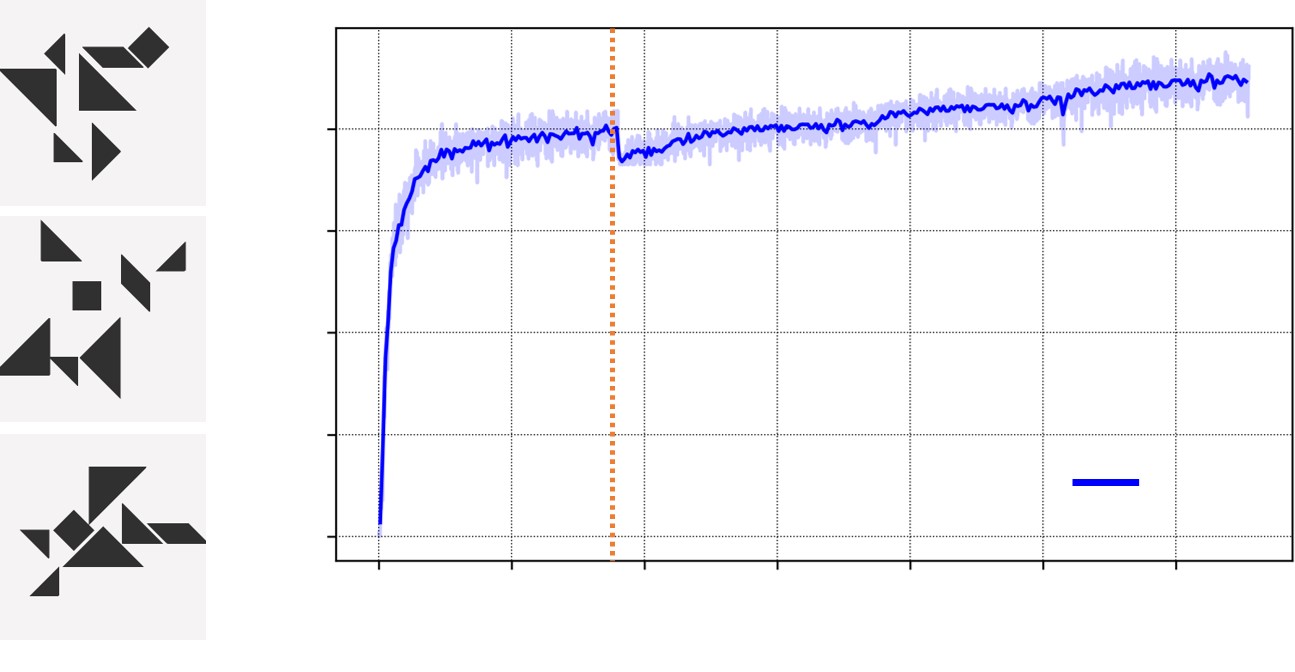}
    \put(6,-0.7) {\footnotesize (a)}
    \put(58,-1.2) {\footnotesize (b)}
    \put(40,2.3) {\footnotesize Training Episodes (Millions)}
    \put(16.5, 10){ {\rotatebox{90}{\footnotesize Relative coverage rate (\%)}}}
    \put(23,9.5) {\footnotesize 0}
    \put(21.5,41) {\footnotesize 80}
    \put(21.5,17.375) {\footnotesize 20}
    \put(21.5,25.25) {\footnotesize 40}
    \put(21.5,33.125) {\footnotesize 60}
    \put(28.5,5.5) {\footnotesize 0}
    \put(88,5.5) {\footnotesize 120}
    \put(38.417,5.5) {\footnotesize 20}
    \put(48.334,5.5) {\footnotesize 40}
    \put(58.251,5.5) {\footnotesize 60}
    \put(68.168,5.5) {\footnotesize 80}
    \put(78,5.5) {\footnotesize 100}
    \put(89,13.5) {\footnotesize PPO}
    \end{overpic}
    \caption{(a) Silhouette examples of randomly generated objects. The top two rows are generated by random placing; the last row is generated with the extra step that adds gravitational force to attract pieces; (b) The relative coverage rate curve of our policy during training. The second stage curriculum begins at the orange line.
    }
    \label{fig:fg4}
\vspace{-0.7cm}
\end{figure}

To learn the tangram assembly, 256 robots in parallel simulation environments collect training episodes using the current policy. In each environment, a random target object, along with an image of a silhouette prompt, is generated. The robot then attempts to predict actions to assemble it according to visual observation and silhouette prompts in the environment, during which the reward is determined by the simulator automatically. If the robot completes the assembly, the environment will be reset, and a new object will be generated again. At last, the collected episodes are returned to the optimizer for learning the policy. We use the Adam optimizer \cite{adam} with a learning rate of $2 \times 10^{-4}$ during the training. Fig. \ref{fig:fg4} shows the learning curves of policy training.

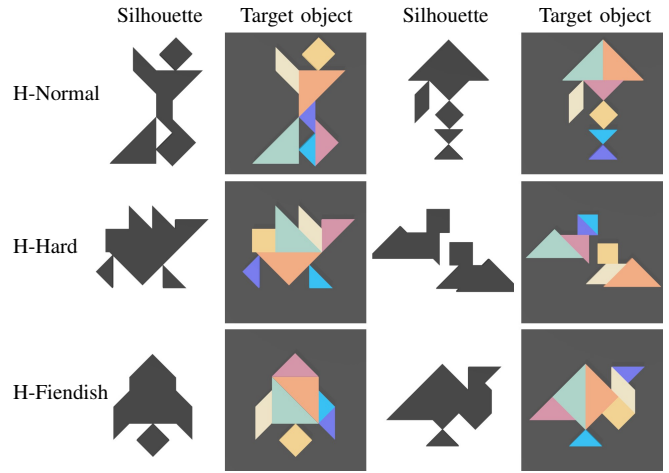
\begin{figure}[!t]
    \centering
    \begin{overpic}[width=\linewidth]{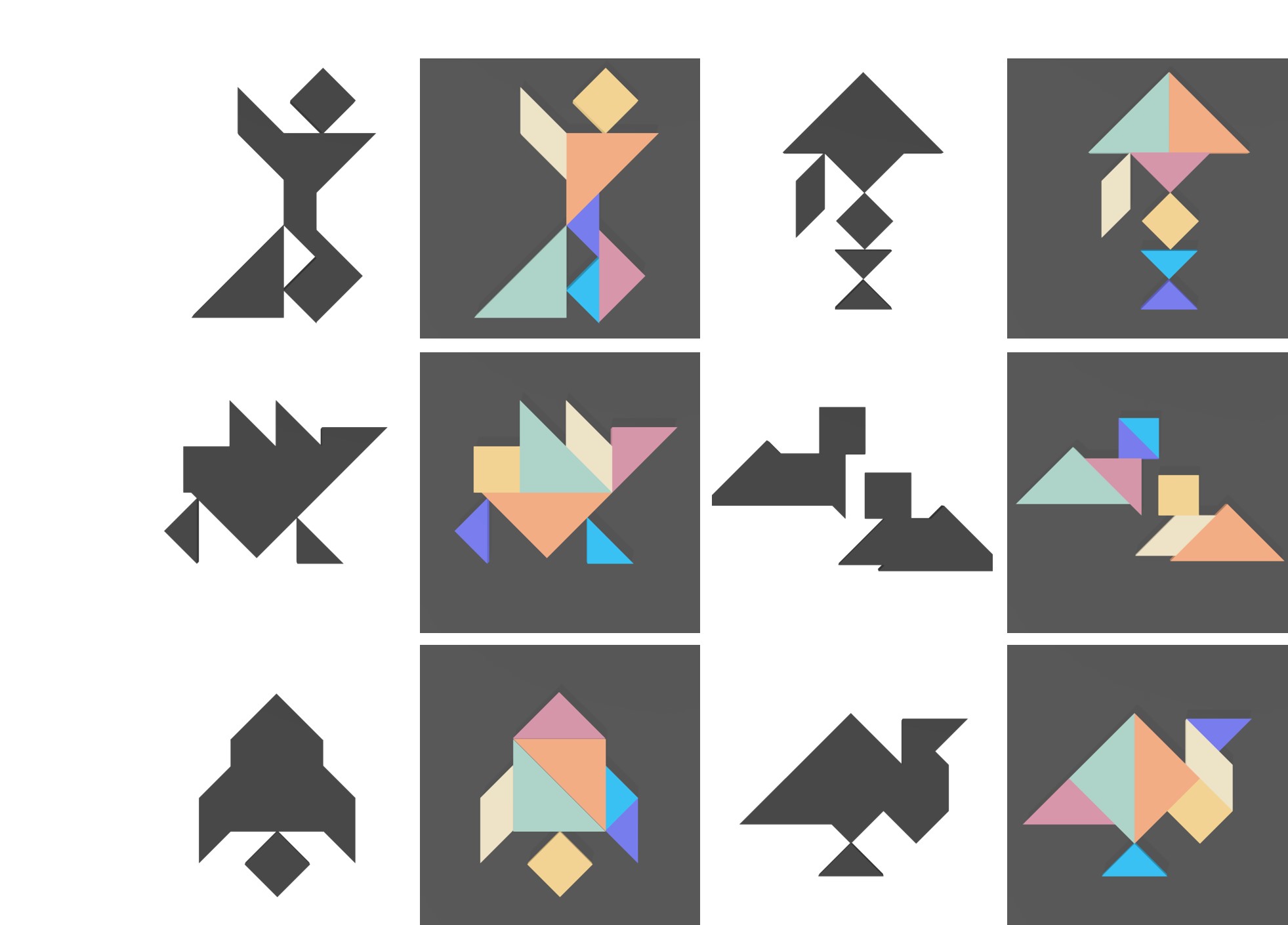}
    \put(0,57) {\footnotesize H-Normal}
    \put(0,33.5) {\footnotesize H-Hard}
    \put(0,11) {\footnotesize H-Fiendish}
    \put(16,69) {\footnotesize Silhouette}
    \put(35,69) {\footnotesize Target object}
    \put(60,69) {\footnotesize Silhouette}
    \put(81,69) {\footnotesize Target object}
    \end{overpic}
    \caption{Example of silhouette prompts and corresponding target objects in different task families. 
    }
    \label{fig:fg5}
\vspace{-0.7cm}
\end{figure}

\section{EXPERIMENTS}\label{sec:exp}

We design a set of experiments to evaluate the effectiveness of MRChaos, in comparison to other baselines across a range of tangram assembly tasks. Additionally, we also investigate an extended version of MRChaos in tasks related to the combination of cutlery and soda. These experiments are conducted in both simulated and real-world settings.

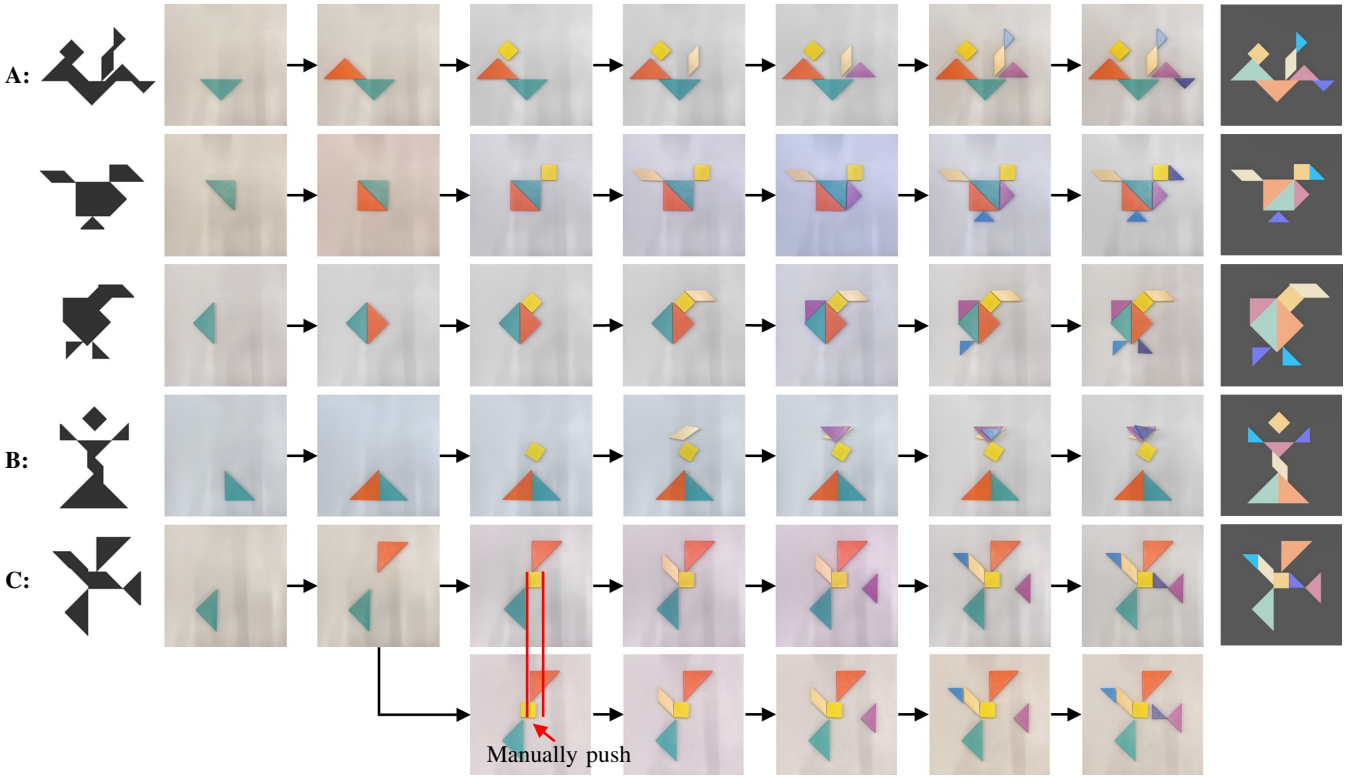
\begin{figure*}[!ht]
    \centering
    \begin{overpic}[width=\linewidth]{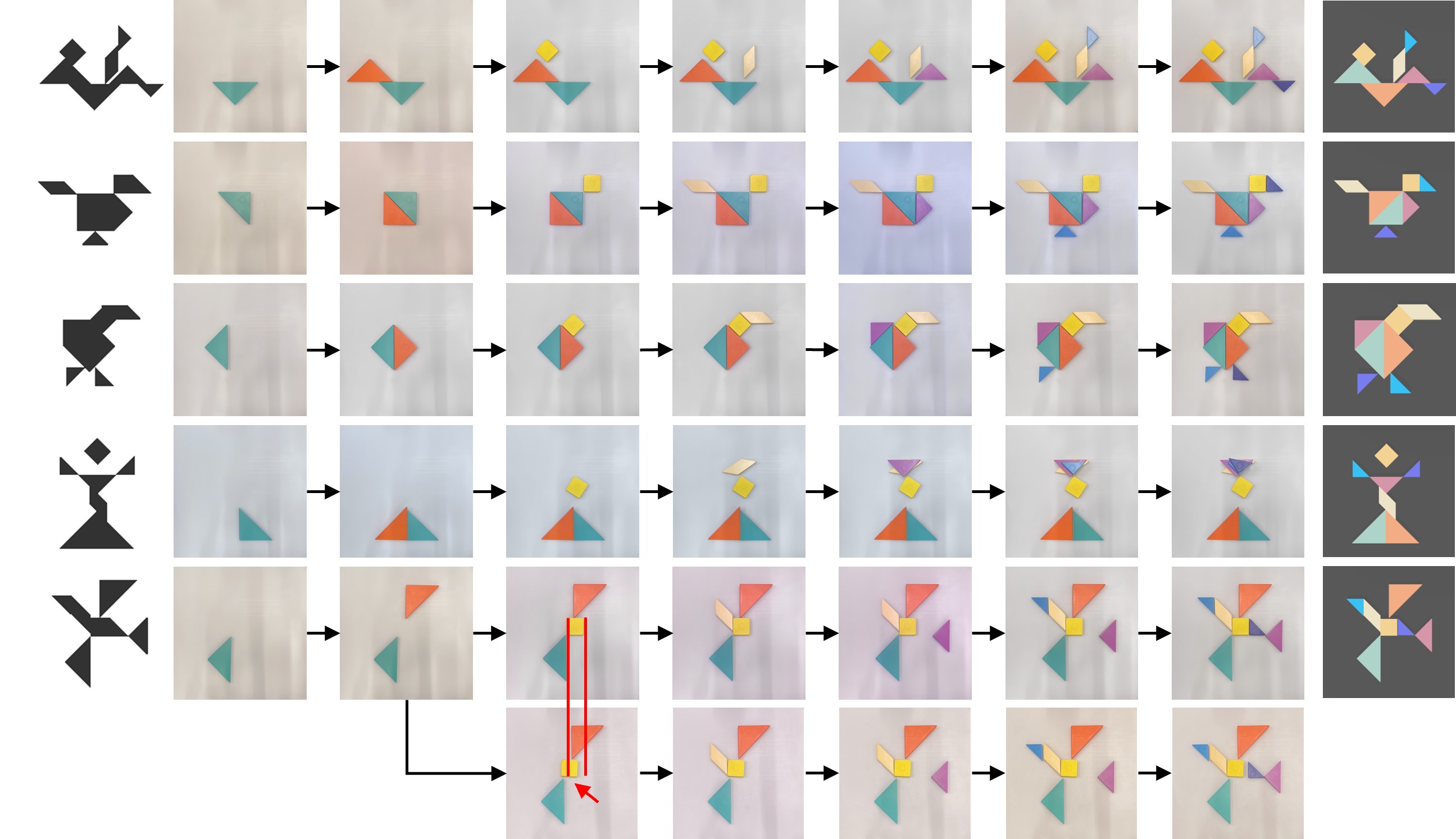}
    \put(36,1) {\small Manually push}
    \put(0,51.75) {\small \textbf{A:}}
    \put(0,23) {\small \textbf{B:}}
    \put(0,14) {\small \textbf{C:}}
    \end{overpic}
    \caption{
    \textbf{Qualitative results of tangram assembly.}
    \textbf{A:} Sequences show MRChaos assembling objects with silhouette prompts. From top to bottom, the objects belong to the H-Normal, H-Hard, and H-Fiendish task families, respectively.
    \textbf{B:} A sequence shows the robot assembling objects through the behavior cloning method;
   \textbf{C:} Sequences demonstrate MRChaos is capable of adapting to the dynamic environment (i.e., the human push the square in the second trial).
    }
    \label{fig:fg6}
\vspace{-0.7cm}
\end{figure*}

\subsection{Evaluation Setup}

\textbf{Evaluation Scene Setup:} We evaluate MRChaos's performance on tangram assembly tasks across four difficulty levels. These different task families cover variations in the complexity of construction, degree of component aggregation, and semantic abstraction. Examples for each level are shown in Fig. \ref{fig:fg5} and fall into the following categories:

\begin{itemize}
\item \textbf{Random:} Target objects generated randomly, using the method outlined in Sec. \ref{sec: method}A, with simple constructions and low component aggregation. These objects contain no semantic information. 

\item \textbf{H-Normal, H-Hard, H-Fiendish:} These target objects within these three task families are created by humans, and each piece is connected to another piece. These families exhibit progressively higher levels of construction complexity and component aggregation. Due to dense aggregation, difficulties arise in precisely locating the pieces and avoiding collisions. Meanwhile, they are highly challenging to construct and may require a high level of imagination and creativity to solve from silhouette. 

\end{itemize}

In particular, objects in H-Normal, H-Hard, and H-Fiendish are previously unseen, while objects in Random are similar to training objects. A total of 104 human-created tangram objects are utilized in the experiments, as shown on our website. They are categorized by the perimeter of the silhouette. Manifestly, the shorter perimeter of the silhouette reflects a higher degree of component aggregation, construction complexity, and reasoning difficulty. 

\textbf{Baseline comparisons:} We compare with the following baseline alternatives:

\begin{itemize}

\item \textbf{\textit{BC}:} A baseline that employs behavior cloning (BC) to directly replicate the perfect assembly policy of 104 human-created tangram objects. Data is collected by disassembling the objects in a simulator, with training data recorded during disassembly.

\item \textbf{\textit{MRChaos w/o local}:} An ablated version of MRChaos that omits the first stage of the training curriculum.

\item \textbf{\textit{MRChaos w/o global}:} An ablated version of MRChaos that omits the second stage of the training curriculum.

\item \textbf{\textit{MRChaos}:} We deploy the MRChaos to the robot, which is the full non-ablated method we propose in this paper. 
\end{itemize}

\textbf{Metric:} We consider two evaluation metrics for validating performance: the relative coverage rate (\textbf{Rela}) and the final coverage rate (\textbf{Final}). The final coverage rate measures the extent to which the assembled object covers the silhouette at the end of the episode. It provides insight into how well the entire assembled object matches the silhouette of the target object, which is useful for evaluating the overall quality of the assembly process. On the other hand, the relative coverage rate is calculated by quantifying the degree of overlap between each placed component and its corresponding area in the silhouette image. This metric is similar to the reward calculation outlined in Sec. \ref{sec: method}D. and provides information on the placement accuracy of each component. 

\begin{table}[!ht]
\centering
\caption{Simulation Results}
\resizebox{\linewidth}{!}{
\begin{tblr}{
  colsep  = 2.5pt,
  rowsep  = 3pt,
  cells = {c},
  cell{1}{1} = {r=2}{},
  cell{1}{2} = {c=2}{},
  cell{1}{4} = {c=2}{},
  cell{1}{6} = {c=2}{},
  cell{1}{8} = {c=2}{},
  hline{1,7} = {-}{},
  hline{3} = {2-9}{},
}
 & Random &  & H-Normal &  & H-Hard &  & H-Fiendish & \\
 & Rela & Final & Rela & Final & Rela & Final & Rela & Final\\
BC & 7.3\% & 18.1\% & 43.8\% & 61.0\% & 40.3\% & 60.6\% & 41.9\% & 62.3\%\\
MRChaos w/o global & 80.2\% & 83.8\% & 53.3\% & 79.4\% & 39.3\% & 70.8\% & 23.8\% & 63.6\%\\
MRChaos w/o local & 49.2\% & 67.6\% & 43.5\% & 68.2\% & 38.3\% & 71.7\% & 34.4\% & 63.9\%\\
\textbf{MRChaos} & \textbf{85.9\%} & \textbf{88.7\%} & \textbf{58.2\%} & \textbf{80.3\%} & \textbf{46.2\%} & \textbf{76.7\%} & \textbf{34.6\%} & \textbf{73.6\%}
\end{tblr}
}
\vspace{-0.65cm}
\label{tab:sim-tangram}
\end{table}

\subsection{Simulation Results}\label{sec: simulation results}

\textbf{Generalization to novel object assembly:} Tab. \ref{tab:sim-tangram} shows the performance of \textit{MRChaos} on different task families in the simulation. The result indicates that \textit{MRChaos} exhibits exceptional abilities in generalization to novel object assembly, despite being trained solely on simple randomly generated objects. In particular, \textit{MRChaos} achieves a final coverage rate of 73.6\% in the H-Fiendish task family, characterized by extreme construction complexity and a high degree of component aggregation. Examples of assembled objects in each task family can be observed in Fig. \ref{fig:fg6}A.

\textbf{Comparison to BC:} From Tab. \ref{tab:sim-tangram}, we can see that the behavior cloning method (\textit{BC}) fails to produce desirable performance despite utilizing human-created tangram objects as its training data. Because \textit{BC} is trained without interaction leads to experiencing a distribution shift and lacks generalizability. As a result, prediction errors accumulate at each step during testing, which limits the ability of behavior cloning to assemble more than the first few pieces correctly. 

    

\textbf{Effectiveness of training curriculum:} We evaluate the effectiveness of the proposed training curriculum by conducting ablation experiments, removing either the first or second stage from \textit{MRChaos}. When comparing \textit{MRChaos} with \textit{MRChaos w/o global}, performance drops by up to 10\% in final coverage rate and 11\% in relative coverage rate on the H-Fiendish task. The results highlight the importance of learning assembly from a global perspective, as accumulated execution errors can affect observations and widen the training-testing domain gap.  Meanwhile, a performance reduction of up to 36\% in relative coverage rate is observed when comparing \textit{MRChaos} to \textit{MRChaos w/o local}, highlighting the difficulty in discovering the correct action during exploration without prior local assembly knowledge.  This shows that allowing the agent first to learn the assembly of local parts facilitates policy learning.

\begin{table}[!t]
\centering
\caption{Real-world Results.}
\resizebox{\linewidth}{!}{
\begin{tblr}{
  cells = {c},
  cell{1}{2} = {c=4}{},
  hline{1,4} = {-}{0.08em},
  hline{3} = {2-5}{0.05em},
}
 & Final coverage rate &  &  & \\
 & Random & H-Simple & H-Middle & H-Hard\\
MRChaos & 83.5\% & 75.2\% & 70.6\% & 62.4\%
\end{tblr}
}
\vspace{-0.65cm}
\label{tab:real-tangram}
\end{table}

\subsection{Real-world Experiments}\label{sec:real_robot_exp}

We also evaluated MRChaos in the real world (Tab. \ref{tab:real-tangram}). Examples of assembled objects in each task family can be observed in Fig. \ref{fig:fg6}A and the accompanying video attachment. The qualitative results in Fig. \ref{fig:fg6}B also show that the performance of BC deteriorates after the initial stages of assembly. This is attributed to the proliferation of prediction errors at each step of the process, which leads to a widening gap between the visual observations in the test and training phases. In contrast, MRChaos manifests robustness to dynamic environments. As shown in Fig. \ref{fig:fg6}C, MRChaos discovers humans modify the position of a piece and plan the remainder of the assembly process in a manner that avoids collision and overlapping. 


\subsection{Extensive Applications of MRChaos}

\begin{figure}[!t]
    \centering
    \begin{overpic}[width=\linewidth]{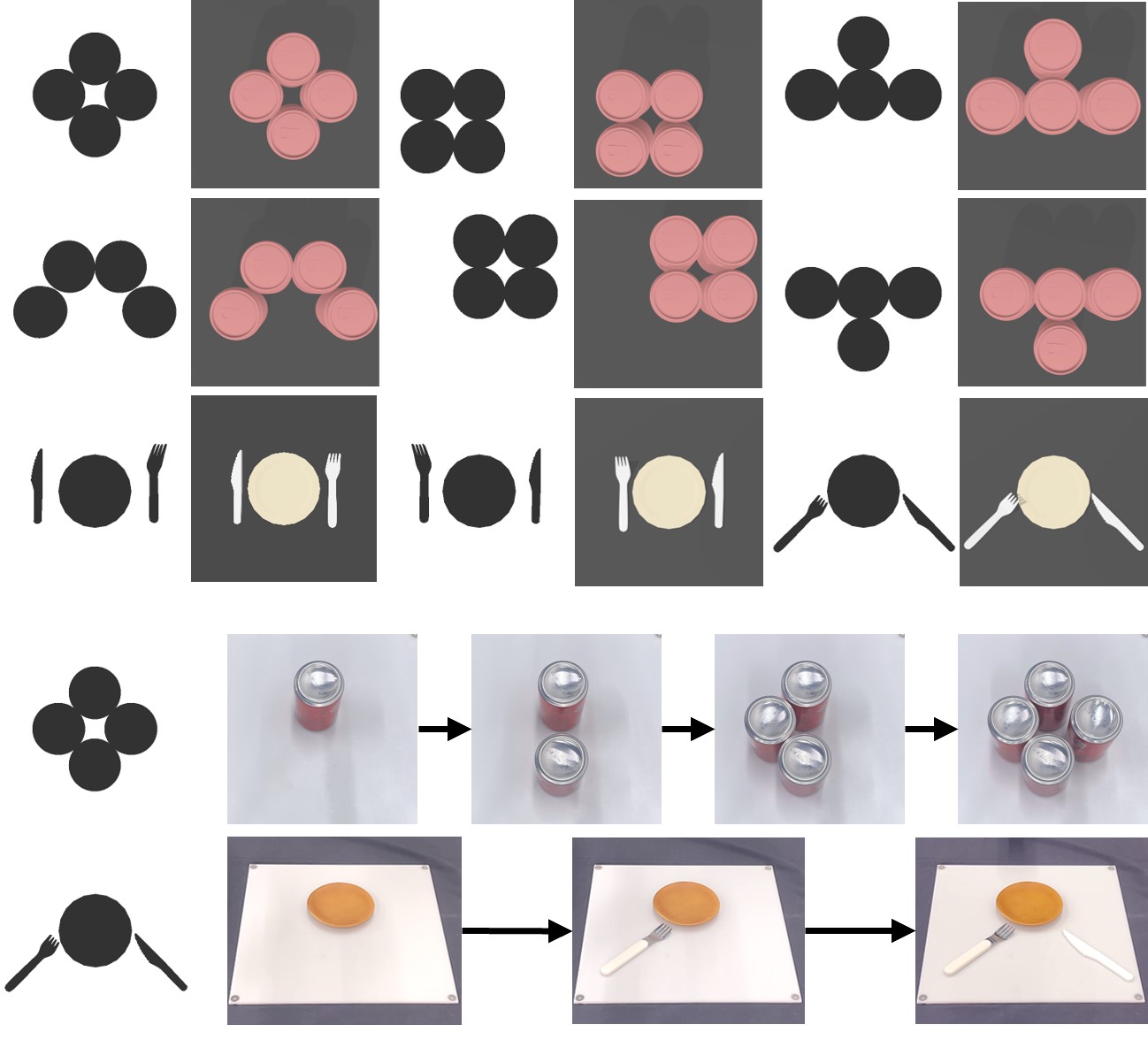}
    \put(48,36.75) {\small (a)}
    \put(48,-1.75) {\small (b)}
    \end{overpic}
    \caption{(a) Silhouettes and corresponding goal configurations of cutlery and soda combination tasks; (b) Sequences of the robot successfully placing the cutlery and soda cans into goal configurations with silhouette prompts.
    }
    \label{fig:fg8}
\vspace{-0.65cm}
\end{figure}

To further demonstrate the generality of the MRChaos, we apply it to two everyday tasks with minimal modifications: a) cutlery combination and b) soda combination. To achieve this, we substitute the tangrams with cutlery and soda cans during training and use the trained model to place objects into different goal configurations as indicated by prompts. The goal configurations and corresponding prompts for these tasks are shown in Fig. \ref{fig:fg8}a, which are previously unseen. We observe that MRChaos successfully placed the cutlery in different semantic targets and soda cans in different combinations (Fig. \ref{fig:fg8}b). On average, MRChaos achieves an 89.7\% final coverage rate for cutlery combination and an 80.6\% final coverage rate for soda combination. For more results, please refer to our video. 

\subsection{Discussion}

Our proposed approach establishes a strong baseline for the tangram assembly task. We have introduced MRChaos, a robust and general approach for learning assembly policies that can generalize to novel objects. Our work promises to open new frontiers in robotic assembly. Prior to our work, a hypothesis could be held that novel object assembly is fundamentally constrained by the complexity of training objects in representing a wide range of real-world objects. Our results indicate that radical generalization in assembly is able to be achieved by learning in much simpler domains without relying on brittle and labor-intensive modeling or costly trial-and-error with real robots. 

We also see a number of limitations and opportunities for future research. First, our task specification of tangram assembly assumes a fixed sequence for piece selection. While the assembly order of tangram pieces can be arbitrary, we hypothesize that planning the order of assembly simultaneously can lead to better assembly results and reduce collisions and stacking. A key direction for future research involves relaxing the strict ordering of pieces, which leads to an important step for robots in terms of mimicking human assembly capabilities. Consequent challenges, such as determining how a robot explores various assembly sequences and assesses their feasibility, could foster methods development. Another hint is that our system utilizes a suction gripper and is limited to planar object rotations. An exciting area for future exploration within tangram assembly involves the integration of more complex manipulation skills using dexterous robotic hands. This could potentially enhance the precision of piece localization and control of contact interactions. 

\section{Conclusion} \label{sec:conclusion}

In this paper, we have presented the first tangram assembly robot and our initial exploration of embodied robotic tangram assembly. We have presented a new approach, MRChaos, for assembling novel objects from silhouette prompts. Our results demonstrate that the intricate nature of object assembly can be effectively tamed without resorting to inflexible and laborious modeling processes or costly and perilous trial-and-error methods in the real world. There are several promising directions for future work. We are particularly interested in investigating humans' low-level manipulation skills with dexterous hands to precisely and reliably place pieces. 




\bibliographystyle{ieeetr}
\bibliography{references}

\end{document}